\documentclass[runningheads]{llncs}
\usepackage[T1]{fontenc}
\usepackage{graphicx,verbatim}
\usepackage{amssymb}
\usepackage{amsmath}
\usepackage{booktabs}
\usepackage{siunitx}
\usepackage{caption}
\usepackage{bm}
\usepackage{booktabs}
\usepackage{hyperref}
\usepackage{xurl}
\usepackage{multirow}
\usepackage{makecell}
\begin{document}
\title{
Robotic Ultrasound Makes CBCT Alive
}
%

\author{Feng Li\inst{1,2}$^{*}$ \and
Ziyuan Li\inst{1}$^{*}$ \and
Zhongliang Jiang\inst{3} \and
Nassir Navab\inst{1,2} \and
Yuan Bi\inst{1,2}}  
\authorrunning{F. Li and Z. Li et al.}
\institute{Chair for Computer-Aided Medical Procedures and Augmented Reality, Technical University of Munich, Munich, Germany \and  Munich Center for Machine Learning, Munich, Germany \and The University of Hong Kong, Hong Kong, China
\\
\email{feng.li@tum.de}}
  
\maketitle              
\begin{abstract}
Intraoperative Cone Beam Computed Tomography (CBCT) provides a reliable 3D anatomical context essential for interventional planning. However, its static nature fails to provide continuous monitoring of soft-tissue deformations induced by respiration, probe pressure, and surgical manipulation, leading to navigation discrepancies. We propose a deformation-aware CBCT updating framework that leverages robotic ultrasound as a dynamic proxy to infer tissue motion and update static CBCT slices in real time. Starting from calibration-initialized alignment with linear correlation of linear combination (LC2)-based rigid refinement, our method establishes accurate multimodal correspondence. To capture intraoperative dynamics, we introduce the ultrasound correlation UNet (USCorUNet), a lightweight network trained with optical flow-guided supervision to learn deformation-aware correlation representations, enabling accurate, real-time dense deformation field estimation from ultrasound streams. The inferred deformation is spatially regularized and transferred to the CBCT reference to produce deformation-consistent visualizations without repeated radiation exposure. We validate the proposed approach through deformation estimation and ultrasound-guided CBCT updating experiments. Results demonstrate real-time end-to-end CBCT slice updating and physically plausible deformation estimation, enabling dynamic refinement of static CBCT guidance during robotic ultrasound-assisted interventions. The source code is publicly available at \url{https://github.com/anonymous-codebase/us-cbct-demo}.


\end{abstract}

\section{Introduction}
In recent years, robotic ultrasound has emerged as a promising paradigm for autonomous and reproducible intraoperative imaging, offering real-time visualization with high soft-tissue contrast and vascular sensitivity \cite{bi2024machine}. Robotic control enables precise probe positioning, regulation of contact force, and automated scanning, supporting applications such as 3D compounding, motion-aware imaging, deformation recovery, and elastography \cite{jiang2023robotic}. However, its limited acoustic window restricts imaging to local soft tissues, which remain prone to artifacts and occlusions. Ultrasound captures only deformation-dependent observations without a consistent global reference, providing dynamic tissue information but not comprehensive anatomical context.

Cone-beam computed tomography (CBCT) provides high-resolution volumetric imaging and has become an important intraoperative modality due to its compact design, lower radiation dose, and flexible integration into the operating room. Commercial systems such as C-arm–based platforms \cite{tanaka2024low} and Loop-X devices \cite{karius2024first} offer on-demand 3D imaging with a large field of view, supporting surgical navigation and serving as detailed volumetric priors. However, CBCT captures only a static snapshot, while anatomy evolves during procedures due to respiration, patient movement, or probe interaction. Repeated CBCT is limited by radiation and workflow constraints, motivating integration with real-time ultrasound. Prior work using electromagnetic \cite{monfardini2021real} and optical tracking \cite{li2025robotic} has enabled rigid CBCT–ultrasound alignment, but deformation-aware multimodal updating remains a key challenge for accurate intraoperative guidance.

In CBCT–ultrasound integration, deformation poses a fundamental challenge \cite{jiang2021deformation}. Ultrasound sequences encode rich temporal information, with adjacent frames reflecting tissue motion from respiration and probe interaction. Classical methods such as speckle tracking \cite{curiale2016influence} and optical flow \cite{li2026ultrasound} have been explored for motion estimation, but modern speckle suppression reduces tracking reliability, and large probe-induced deformations remain difficult to capture. Deep learning–based optical flow models, like recurrent all-pairs field transforms (RAFT) \cite{teed2020raft}, improve robustness to large displacements, yet ultrasound artifacts and depth-dependent distortions violate standard assumptions. Purely data-driven approaches also lack physical constraints, potentially producing biomechanically implausible deformation fields. These limitations motivate dedicated deformation modeling for CBCT–ultrasound integration.

To address these challenges, we propose a robotic-ultrasound-driven framework for deformation-aware CBCT updating in image-guided intervention. Our main contributions are: (1) a workflow-compatible CBCT--ultrasound pipeline integrating calibration, LC2 refinement, ultrasound deformation estimation, and ultrasound-informed CBCT slice update; (2) USCorUNet, a lightweight bidirectional correlation-enhanced network with optical flow-guided training; (3) real-time CBCT slice updating for probe- and externally induced deformations; and (4) multi-dataset in vivo and phantom validation, showing a favorable accuracy--efficiency trade-off against RAFT-based and classical baselines.

\section{Methods}

\begin{figure}[t]
    \centering
    \includegraphics[width=\linewidth]{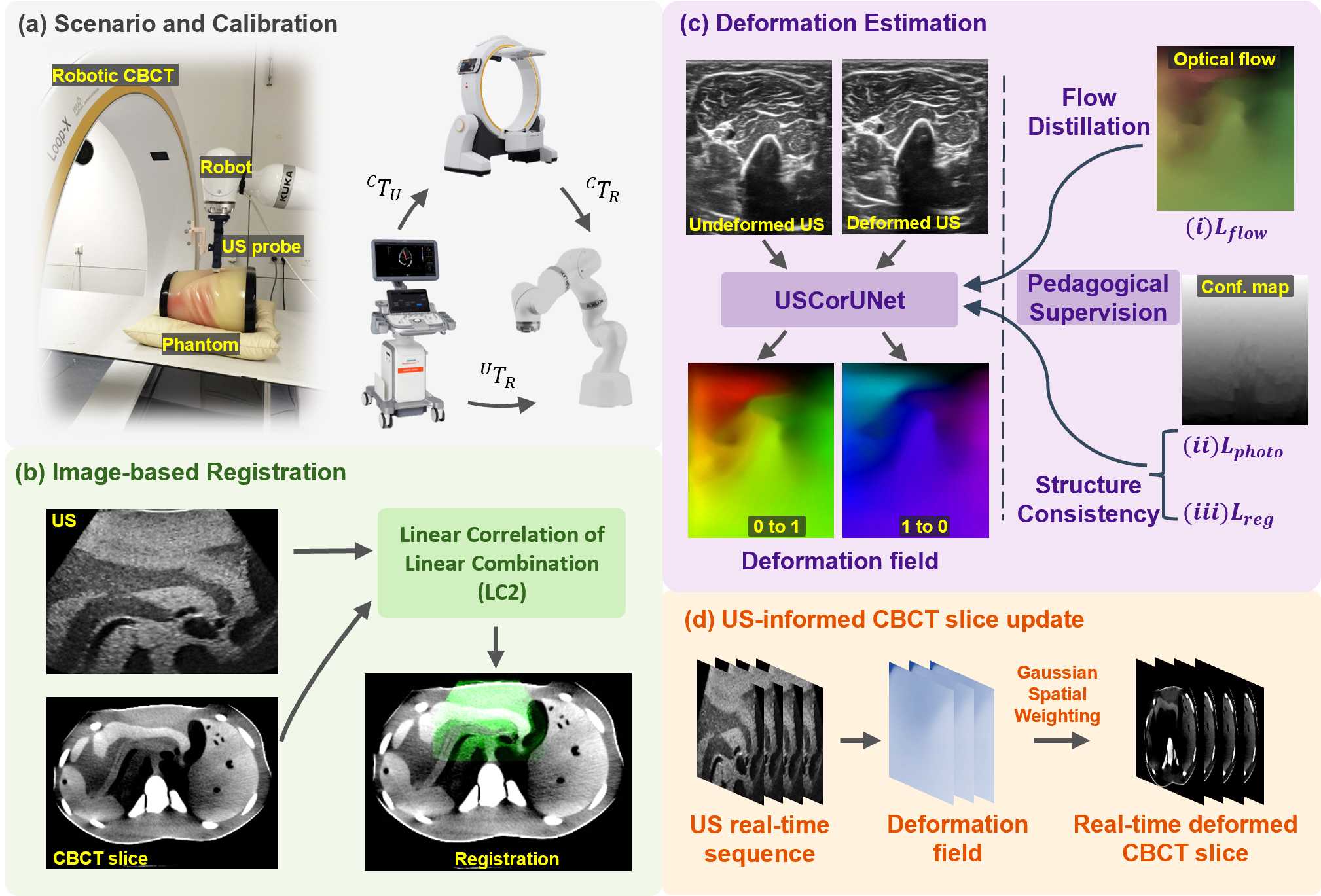}
    \caption{System overview. (a) Rigid calibration between robotic ultrasound and CBCT. (b) Image-based registration refinement. (c) Deformation estimation with USCorUNet. (d) Deformation transfer for CBCT slice updating. Ultrasound examples from the in vivo arm dataset (c) and the CT-mapped phantom dataset (b,d) illustrate cross-domain applicability. Conf. map denotes confidence map.}
    \label{fig:pipeline}
\end{figure}

Our system has four modules (Fig.~\ref{fig:pipeline}). Calibration-based rigid initialization establishes the CBCT–ultrasound spatial relationship (Fig.~\ref{fig:pipeline}(a)), refined by LC2 to correct residual alignment errors (Fig.~\ref{fig:pipeline}(b)). Non-rigid deformation between consecutive ultrasound frames is estimated with USCorUNet using pedagogical supervision (Fig.~\ref{fig:pipeline}(c)). The resulting deformation is applied to update the corresponding CBCT slice, producing a dynamically deformed CBCT for visualization and guidance (Fig.~\ref{fig:pipeline}(d)).

\subsection{Calibration and Registration}\label{sec:Registration}
For spatial calibration, we adopted the hand-eye calibration framework proposed in \cite{li2025robotic}. Ultrasound (US) images can be projected into the CBCT volume via the spatial relationships among the CBCT, robot, and probe. The rigid transformation between a US image and CBCT slice, $^{C}\mathbf{T}_{U}$, is computed as $^{C}\mathbf {T}_{U} = ^{C}\mathbf {T}_{R}(^{U}\mathbf {T}_{R})^{-1}$, where $^{C}\mathbf {T}_{R}$ and $^{U}\mathbf {T}_{R}$ map the CBCT and US image to the robot base frame.

Although calibration achieves a mean alignment error of approximately 1 - 2 mm, residual misalignment persists without image-based refinement, which can be clinically significant in precision-sensitive procedures such as needle insertion. To improve accuracy, we incorporate multimodal rigid registration using the LC2 similarity metric \cite{wein2013global}, which models the relationship between CBCT intensities and ultrasound appearance via a local linear approximation: $f(x_i) = \alpha \mathbf{p_i} + \beta \mathbf{g_i} + \gamma$, where $p_i$ and $g_i$ denote CBCT intensity and gradient magnitude, and ${\alpha, \beta, \gamma}$ estimate the local cross-modal relationship. Initialized by calibration, LC2 searches within a constrained range, reducing computational cost and runtime.
 
\subsection{Deformation Field Acquisition}\label{sec:Deformation}
Building on the rigid alignment, USCorUNet estimates dense bidirectional deformation fields between ultrasound frames $I_0, I_1 \in \mathbb{R}^{H\times W}$, yielding $F_{01}, F_{10}\in\mathbb{R}^{H\times W\times 2}$. We distill pseudo-labels from RAFT~\cite{teed2020raft}. For each pair, we generate a direct RAFT candidate on $(I_0,I_1)$ and a bisect candidate via an intermediate frame, composed as $(F\oplus G)(\mathbf{x})=F(\mathbf{x})+G(\mathbf{x}+F(\mathbf{x}))$, and select the candidate with lower post-warp misalignment under differentiable warping $\mathcal{W}(I,F)(\mathbf{x})=I(\mathbf{x}+F(\mathbf{x}))$.

\begin{figure}
    \centering
    \includegraphics[width=0.9\linewidth]{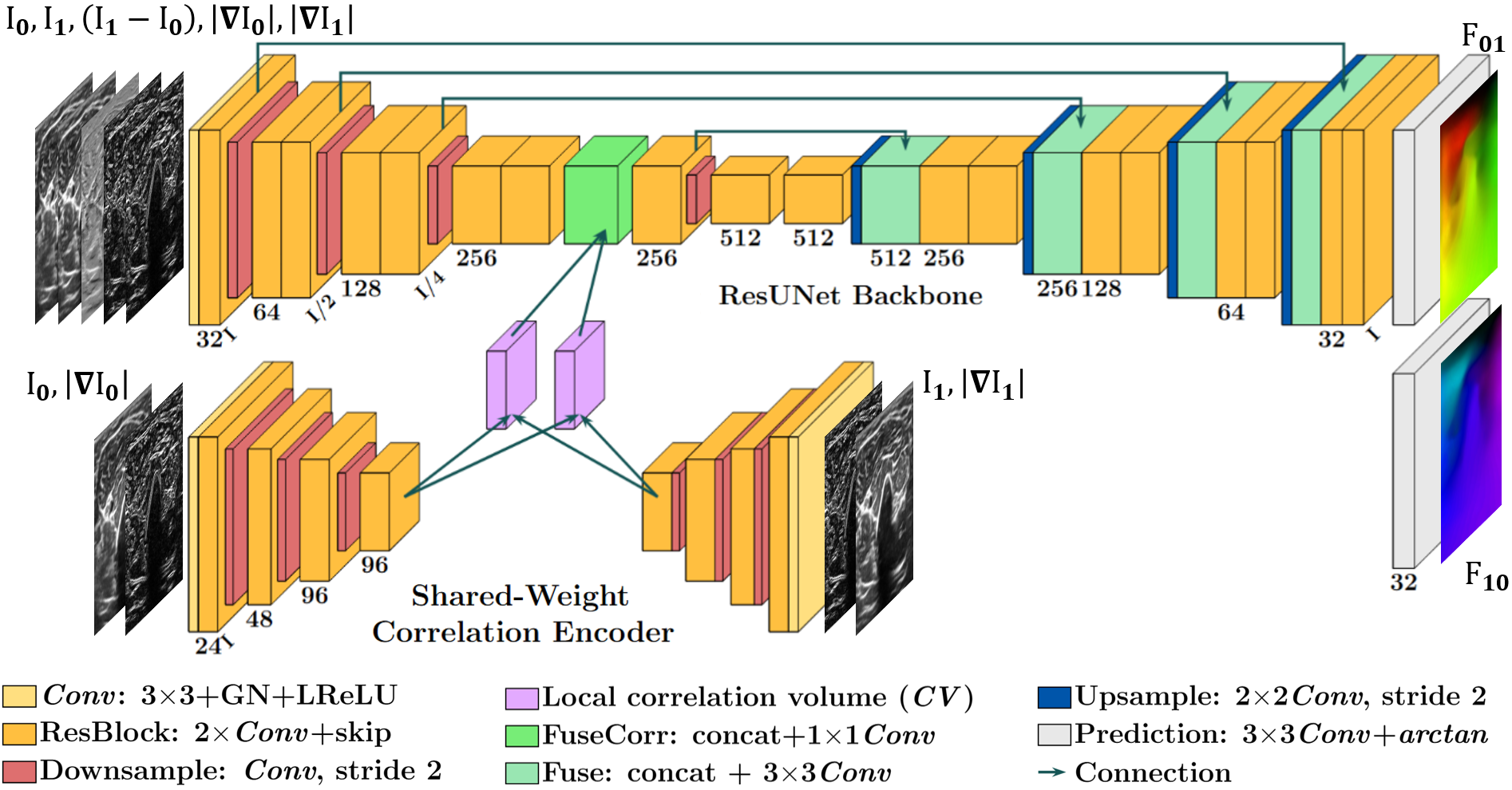}
    \caption{Architecture of USCorUNet.}
    \label{fig:USCorUNet}
\end{figure}
Fig.~\ref{fig:USCorUNet} shows the architecture of USCorUNet. A preprocessing module forms a five-channel input $\big[I_0, I_1, (I_1-I_0), |\nabla I_0|, |\nabla I_1|\big]$. The network combines a ResUNet-style context encoder-decoder with a shared-weight correlation encoder $g_{\phi}$. From features $f_i=g_{\phi}([I_i,|\nabla I_i|])$, we build local correlation volumes, e.g., $CV_{01}(\mathbf{x},\Delta)=\frac{1}{\sqrt{C}}\langle f_0(\mathbf{x}), f_1(\mathbf{x}+\Delta)\rangle$, and define $CV_{10}$ analogously, where $C$ is the feature dimension and $\Delta$ indexes a local displacement neighborhood. The correlation volumes are fused with same-scale context features and decoded into dense fields at $1/8$ resolution, balancing efficiency and reliable matching.


Since appearance-driven pseudo-labels do not explicitly enforce physical plausibility~\cite{greer2021icon}, we train USCorUNet with a bidirectional objective that combines optical flow distillation $\mathcal{L}_{\text{flow}}$, confidence-weighted photometric consistency $\mathcal{L}_{\text{photo}}$, and regularization $\mathcal{L}_{\text{reg}}$. Specifically, $\mathcal{L}_{\text{flow}}$ is an $\ell_1$ loss to the optical flows, $\mathcal{L}_{\text{photo}}$ is a confidence-weighted Charbonnier penalty~\cite{bruhn2005lucas} on post-warp intensity residuals using random-walk confidence maps~\cite{karamalis2012ultrasound}, and $\mathcal{L}_{\text{reg}}$ combines edge-aware smoothness~\cite{janai2018unsupervised} and a Jacobian-based folding penalty~\cite{shen2019networks}. The final objective is $\mathcal{L}=\lambda_{\text{flow}}\mathcal{L}_{\text{flow}}+\lambda_{\text{photo}}\mathcal{L}_{\text{photo}}+\lambda_{\text{reg}}\mathcal{L}_{\text{reg}}$, with $\lambda_{\text{flow}}=1$, $\lambda_{\text{photo}}=0.2$, and $\lambda_{\text{reg}}=0.05$, chosen to prioritize flow distillation while using photometric consistency as a complementary cue and regularization as a mild physical prior. 

\subsection{Ultrasound Guided CBCT Updating}\label{sec:CBCT}
The estimated deformation field updates the CBCT slice in real time, accounting for probe and external motion. Specifically for probe-induced motion, we correct non-uniform convex compression using a Gaussian profile
$P(x) = d_{\text{robot}} \cdot \exp( - (x - c_x)^2 / 2\sigma_{\text{probe}}^2 )$,
where $d_{\text{robot}}$ denotes the probe displacement magnitude, $c_x$ the lateral midpoint, and $\sigma_{\text{probe}}$ the curvature. This profile corrects the vertical deformation component: $\mathbf{D}{\text{geo}}^y=\mathbf{D}{\text{raw}}^y-P(x)$.


To align the local ultrasound field with the larger CBCT ROI, we use Euclidean Distance Transform (EDT)-based spatial weighting. The corrected field is padded to $\mathbf{D}{\text{pad}}$ and scaled by $W(x,y)=\exp(-\mathcal{D}(x,y)/\sigma{\text{smooth}})$, where $\mathcal{D}=\text{EDT}(1-M)$ denotes the distance to the ultrasound boundary. The final field $\mathbf{D}{\text{final}}=\mathbf{D}{\text{pad}}\odot W$ enforces smooth decay of deformation from the probe contact region across the CBCT slice.

\section{Experiments and Results}
\subsection{Setup, Datasets, and Experimental Details}



The experimental setup integrates a KUKA LBR iiwa 14 R820 robot and Siemens ACUSON Juniper ultrasound (5C1 probe) via a 3D-printed holder, with images acquired through an Epiphan frame grabber. A Loop-X Imaging Ring provided CBCT data.

Experiments were conducted on four datasets: (A) in vivo forearm/upper-arm ultrasound; (B) a pork-tissue gel phantom; (C) a chicken/pork gel phantom; and (D) an abdominal phantom (Kyoto Kagaku US-22). Dataset B additionally includes finger-press compression to simulate externally induced deformation. 

USCorUNet was first trained on Dataset~A (8:1:1 split) to obtain a base model, using $(I_0, I_1)$ pairs and AdamW for 50 epochs (batch size 4, learning rate $2\times10^{-4}$, weight decay $10^{-4}$, mixed precision). Starting from this base model, we fine-tuned two regime-specific models: (i) a model fine-tuned for probe-induced motion on Datasets~B--D, and (ii) a model fine-tuned for externally induced motion on Dataset~B. Each variant was fine-tuned for 20 epochs with a learning rate of $10^{-4}$; the data split and all other training settings were kept unchanged.



\subsection{Metrics, Baselines, and Ablations}
We evaluate bidirectional deformation estimates using (i) post-warp alignment (MAE, NCC), (ii) forward--backward (FB) consistency via the mean $\ell_2$ norm of $r_{01}=F_{01}\oplus F_{10}$ and $r_{10}=F_{10}\oplus F_{01}$ (mean FB residual)~\cite{varadhan2013framework}, and (iii) physical plausibility via the folding ratio (fraction of pixels with $\det(I+\nabla F)<0$). On Dataset~A, we additionally report Dice on SAM-segmented~\cite{kirillov2023segment} bone masks after nearest-neighbor warping. For deformation-warped CT volumes, we also report bone-mask Dice and SSIM.

As the primary baseline, we evaluate the selected optical flow under the same metrics as a reference for USCorUNet. We further conduct Dataset~A ablations to assess key components: (i) removing the correlation branch (w/o Corr.), (ii) disabling confidence-map weighting (w/o Conf.), and (iii) simplifying the training objective to $\mathcal{L}_{\text{flow}}+\mathcal{L}_{\text{photo}}$ or $\mathcal{L}_{\text{flow}}+\mathcal{L}_{\text{reg}}$.



\subsection{Results of Deformation Field Acquisition}
\begin{figure}
    \centering
    \includegraphics[width=0.9\linewidth]{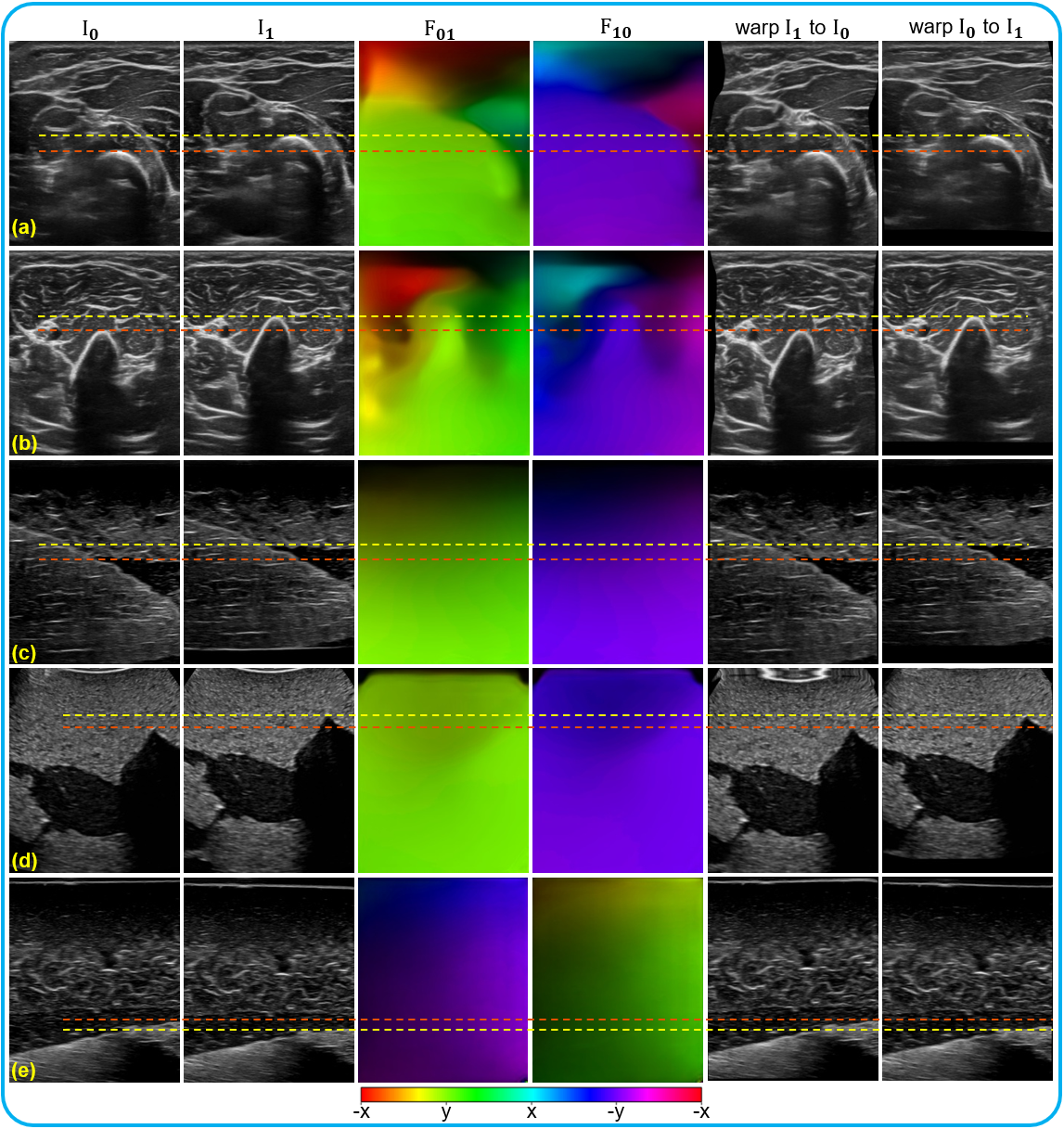}
    \caption{Bidirectional deformation estimation results. (a,b) In vivo arm examples (Dataset~A); (c,d) probe-induced motion (Datasets~B and~D); (e) externally induced motion (Dataset~B). Orange/yellow dashed lines indicate visual alignment guides for $I_0$, $I_1$. The color bar indicates flow direction (x/y).}
    \label{fig:deformation_field}
\end{figure}

\subsubsection{Base model.}
Table~\ref{tab:results_stage1_ablation} shows that USCorUNet matches the optical flow in photometric alignment, slightly improves bone-mask Dice, and markedly improves deformation quality, reducing the mean FB residual by around 53\% and lowering the folding ratio. Ablations further show that removing the correlation branch causes the largest drop in photometric alignment with only a mild Dice decrease, suggesting its primary benefit is improved correspondence in non-bone texture regions. Disabling confidence weighting or removing \(\mathcal{L}_{\text{reg}}\) mainly worsens FB consistency and increases foldings, whereas removing \(\mathcal{L}_{\text{photo}}\) yields the largest Dice drop, indicating that photometric supervision is important for anatomically faithful deformations.

\begin{table}
\centering
\caption{Testset performance of the base model on Datasets~A, averaged over both deformation directions ($I_0\!\rightarrow\!I_1$, $I_1\!\rightarrow\!I_0$) and reported as mean $\pm$ standard deviation. $\downarrow/\uparrow$ indicate lower/higher is better. Best results are in bold.}
\label{tab:results_stage1_ablation}
\setlength{\tabcolsep}{2pt}
\footnotesize
\begin{tabular}{lccccc}
\hline
Method & MAE $\downarrow$ & NCC $\uparrow$ & FB $\downarrow$ & Fold(\% ) $\downarrow$ & Dice(\% ) $\uparrow$ \\
\hline
\multicolumn{6}{l}{\textit{Baseline}} \\
RAFT & \bm{$0.05 \pm 0.02$} & \bm{$0.85 \pm 0.10$} & $1.81 \pm 1.22$ & $0.24 \pm 0.14$ & $90.31 \pm 4.14$ \\
\hline
\multicolumn{6}{l}{\textit{Ours}} \\
USCorUNet& \bm{$0.05 \pm 0.02 $} & \bm{$0.85 \pm 0.09$} & \bm{$0.85 \pm 0.57$} & \bm{$0.13 \pm 0.10$} & \bm{$90.62 \pm 3.76$} \\
\hline
\multicolumn{6}{l}{\textit{Ablation Study}} \\
w/o Corr. & $0.08 \pm 0.03$ & $0.61 \pm 0.18$ & $1.32 \pm 3.14$ & $0.56 \pm 0.51$ & $89.10 \pm 3.33$ \\
w/o Conf. & $0.06 \pm 0.06$ & $0.74 \pm 0.16$ & $1.55 \pm 1.20$ & $0.17 \pm 0.11$ & $87.03 \pm 6.17$ \\
w/o $\mathcal{L}_{\text{reg}}$ & $0.06 \pm 0.02$ & $0.79 \pm 0.12$ & $1.21 \pm 0.92$ & $0.41 \pm 0.22$ & $89.30 \pm 4.23$ \\
w/o $\mathcal{L}_{\text{photo}}$ & $0.07 \pm 0.03$ & $0.72 \pm 0.17$ & $1.38 \pm 1.03$ & $0.19 \pm 0.19$ & $85.45 \pm 6.28$ \\
\hline
\end{tabular}
\end{table}

Table~\ref{tab:dice_compare_defcornet} compares USCorUNet with DefCor-Net~\cite{jiang2023defcor} on the same in-vivo ultrasound dataset (Dataset~A) under their force-stratified evaluation protocol. Improvements are more pronounced at higher forces, which correspond to larger and more challenging compressions.
\begin{table}
\centering
\caption{Bone-mask Dice under the force-stratified protocol of DefCor-Net~\cite{jiang2023defcor} on Dataset~A. Dice is reported for the $I_1\!\rightarrow\!I_0$ direction.}
\label{tab:dice_compare_defcornet}
\setlength{\tabcolsep}{1.5pt}
\begin{tabular}{lcccccc}
\hline
Method & 1\,N & 2\,N & 3\,N & 4\,N & 5\,N & 6\,N \\
\hline
DefCor-Net & \bm{$95.9\pm3.3$} & $92.4\pm4.8$ & $91.1\pm7.5$ & $87.8\pm7.1$ & $87.8\pm12.2$ & $82.6\pm12.1$ \\
USCorUNet & $95.4\pm2.6$ & \bm{$94.1\pm2.4$} & \bm{$94.0\pm3.6$} & \bm{$93.2\pm1.8$} & \bm{$89.8\pm1.5$} & \bm{$87.5\pm1.2$} \\
\hline
\end{tabular}
\end{table}

\subsubsection{Fine-tuned models.}

Table~\ref{tab:results_finetune} shows that regime-specific fine-tuning improves performance, particularly FB consistency and physical plausibility. For probe-induced motion, it reduces mean FB residuals and folding ratios while maintaining alignment. For externally induced motion, it mitigates domain shift and restores alignment. Overall, the base checkpoint provides a robust and transferable initialization. Visual results are shown in Fig.~\ref{fig:deformation_field}.

\begin{table}
\centering
\caption{Test-set performance of fine-tuned models on Datasets~B--D (probe-induced) and Dataset~B (externally induced), averaged over both directions.}
\label{tab:results_finetune}
\setlength{\tabcolsep}{2.5pt}
\begin{tabular}{l|lcccc}
\hline
Motion & Method & MAE $\downarrow$ & NCC $\uparrow$ & FB $\downarrow$ & Fold(\% ) $\downarrow$ \\
\hline
\multirow{3}{*}{\makecell[l]{Probe- \\ induced}}
&RAFT & \bm{$0.03 \pm 0.01$} & \bm{$0.94 \pm 0.03$} & $1.03 \pm 2.27$ & $0.23 \pm 0.14$ \\
&Base model & $0.04 \pm 0.01$ & $0.84 \pm 0.08$ & $0.84 \pm 0.65$ & $0.15 \pm 0.11$ \\
&Probe-adapted & \bm{$0.03 \pm 0.01$} & $0.93 \pm 0.03$ & \bm{$0.33 \pm 0.28$} & \bm{$0.09 \pm 0.10$} \\
\hline
\multirow{3}{*}{\makecell[l]{Externally \\ induced}}
&RAFT & \bm{$0.03 \pm 0.01$} & \bm{$0.93 \pm 0.05$} & $0.91 \pm 1.35$ & $0.14 \pm 0.13$ \\
&Base model & $0.05 \pm 0.02$ & $0.71 \pm 0.17$ & $1.28 \pm 1.95$ & $0.45 \pm 0.55$ \\
&External-adapted & \bm{$0.03 \pm 0.01$} & $0.92 \pm 0.05$ & \bm{$0.23 \pm 0.24$} & \bm{$0.07 \pm 0.07$} \\
\hline
\end{tabular}
\end{table}

\subsection{Results of Ultrasound Guided CBCT Updating}
\begin{figure}
    \centering
    \includegraphics[width=0.9\linewidth]{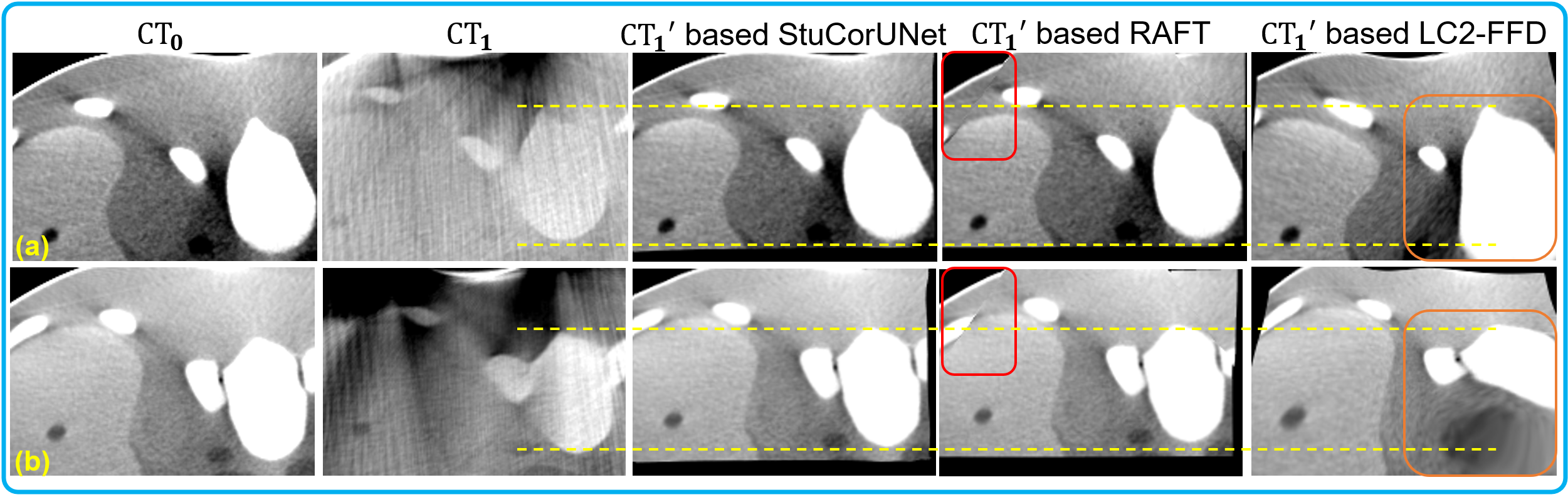}
    \caption{$CT_1'$ update results on two representative abdominal phantom cases (Dataset~D) using USCorUNet, RAFT, and LC2-FFD ($CT_0$: source; $CT_1$: target). Red/orange boxes indicate structural artifacts/severe deformation regions.}
    \label{fig:cbct_updating}
\end{figure}


Table~\ref{tab:cbct_updating_results} compares our method with RAFT-based and classical LC2-FFD \cite{fuerst2014automatic} baselines. Our approach achieves the best quality–efficiency trade-off, slightly outperforming RAFT while reducing runtime by $5\times$. In addition, RAFT can exhibit tearing artifacts. Although LC2-FFD shows comparable MAE and SSIM, it introduces geometric distortions and is $512\times$ slower. As absolute metrics are affected by robotic artifacts, this serves as a controlled relative comparison. Visual results are shown in Fig.~\ref{fig:cbct_updating}.

\begin{table}
\centering
\caption{Quantitative results for CBCT updating on Dataset~D using ultrasound-based deformation estimation, with all three methods evaluated under identical conditions and runtime reported end-to-end.}
\label{tab:cbct_updating_results}
\setlength{\tabcolsep}{4pt}
\begin{tabular}{lcccc}
\hline
Model & MAE $\downarrow$ & SSIM $\uparrow$ & Dice (\%) $\uparrow$ & Timing (ms) $\downarrow$ \\
\hline
USCorUNet & \textbf{0.33 $\pm$ 0.04} & \textbf{0.22 $\pm$ 0.02} & \textbf{82.22 $\pm$ 10.54} & \textbf{11.25 $\pm$ 0.44} \\
RAFT & 0.34 $\pm$ 0.04 & 0.21 $\pm$ 0.03 & 79.86 $\pm$ 12.31 & 56.24 $\pm$ 1.78 \\
LC2-FFD & \textbf{0.33 $\pm$ 0.05} & \textbf{0.22 $\pm$ 0.03}& 58.91 $\pm$ 8.77 & 5764.26 $\pm$ 510.63 \\
\hline
\end{tabular}
\end{table}

\section{Discussion and Conclusion}\label{sec:Conclusion}
In this paper, we introduced a deformation-aware CBCT updating framework that integrates rigid calibration, registration, and deformation field estimation for real-time end-to-end CBCT slice updating and enhanced intraoperative visualization. The proposed USCorUNet efficiently estimates deformation fields from adjacent ultrasound frames while preserving structural consistency.

Compared to RAFT and LC2-FFD baselines, our method achieves a superior trade-off between registration accuracy and computational efficiency. A promising avenue for future work involves incorporating semantic segmentation into the registration pipeline. While the current approach relies on intensity and structural features, integrating semantic information could further refine deformation details in complex anatomical regions, ultimately paving the way for more reliable intraoperative guidance.

%
%
%
\clearpage
\bibliographystyle{splncs04}
\bibliography{bibliography}

\end{document}